# DeepCCI: End-to-end Deep Learning for Chemical-Chemical Interaction Prediction

Sunyoung Kwon
Electrical and Computer Engineering
Seoul National University
Seoul 08826, Korea

Sungroh Yoon*
Electrical and Computer Engineering
Seoul National University
Seoul 08826, Korea
sryoon@snu.ac.kr

## ABSTRACT

Chemical-chemical interaction (CCI) plays a key role in predicting candidate drugs, toxicity, therapeutic effects, and biological functions. In various types of chemical analyses, computational approaches are often required due to the amount of data that needs to be handled. The recent remarkable growth and outstanding performance of deep learning have attracted considerable research attention. However, even in state-of-the-art drug analysis methods, deep learning continues to be used only as a classifier, although deep learning is capable of not only simple classification but also automated feature extraction. In this paper, we propose the first end-to-end learning method for CCI, named DeepCCI. Hidden features are derived from a simplified molecular input line entry system (SMILES), which is a string notation representing the chemical structure, instead of learning from crafted features. To discover hidden representations for the SMILES strings, we use convolutional neural networks (CNNs). To guarantee the commutative property for homogeneous interaction, we apply model sharing and hidden representation merging techniques. The performance of DeepCCI was compared with a plain deep classifier and conventional machine learning methods. The proposed DeepCCI showed the best performance in all seven evaluation metrics used. In addition, the commutative property was experimentally validated. The automatically extracted features through end-to-end SMILES learning alleviates the significant efforts required for manual feature engineering. It is expected to improve prediction performance in drug analyses.

## GENERAL TERMS

Algorithms

*To whom correspondence should be addressed.

## 1 INTRODUCTION

Interaction is an action that occurs between two entities that may share similar functions or metabolic pathways [4, 41, 50]. Various interactions exist, such as protein-protein [15, 24, 47], compound-protein [58], miRNA-mRNA [33], and chemical-chemical interactions (CCI) [31].

Previous studies on computational approaches to CCI include the following: a study on compound synergism that predicts the therapeutic efficacy of molecule combinations, based on interactive chemicals [66]; toxicity related studies on predicting chemical toxicities and side effects based on the assumption that interacting chemicals are more likely to share a similar toxicity [6, 7]; targeted candidate drug discovery studies through which the new closest candidate drug to the commercialized target drug was proposed by connecting the interacting chemicals in the graphical aspects [8]; and a study in which a novel approach was developed for identifying biological functions of chemicals, based on the assumption that interactive chemicals participate in the same metabolic pathways [22]. As evidenced by the range of studies mentioned above, CCI has been used for many purposes. However, learning-based methods for CCI are hard to find. We therefore propose a learning-based CCI prediction method in this paper.

The conventional learning-based chemical (drug) analyses can be divided into two stages, feature engineering and predicting activity, as shown in Figure 1. In drug analyses, features are also called descriptors, fingerprints, or molecular representations.

The feature engineering stage extracts informative features from chemicals. A significant amount of information has been manually designed by domain experts, ranging from simple counts (*e.g.*, the number of atoms, rings, and bonds) to topological properties (*e.g.*, atom pairs, shapes, and connectivities) [59]. Extracted features have been used in subsequent analyses and significantly affected prediction performance. Various works have been proposed to capture the essential properties of chemicals, and more than 5,000 diverse features have been identified [59] (*e.g.*, PubChem, MACCS, and CDK fingerprint).

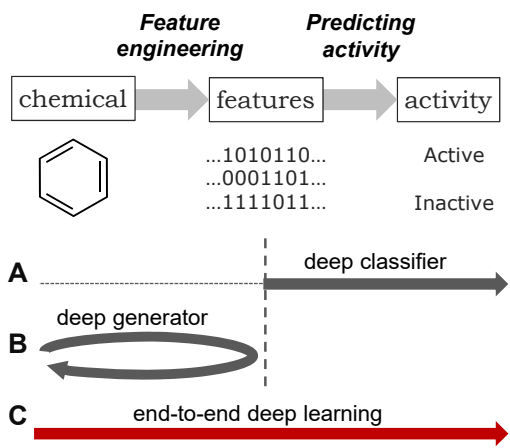

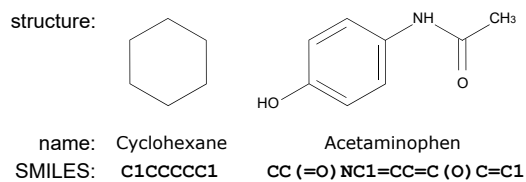

Figure 2: Examples of chemical compounds: 2D structure, name, and SMILES

Figure 1: Two-stage chemical analysis: feature engineering and predicting activity. Three types of deep learning approaches are shown: (A) deep classifier, whereby deep learning is applied to the back-end, (B) deep generator, whereby deep learning is applied to the front-end, and (C) end-to-end deep learning, whereby deep learning is applied from the front-end to back-end (our approach).

The predicting activity stage uncovers the relationship between the extracted features and biological activities, such as absorption, distribution, metabolism, excretion, and toxicity (ADMET) [57, 60]. To robustly learn relationships between chemicals and activities, and to handle a large amount of data, machine learning approaches (such as support vector machine (SVM) [10, 13] and random forest (RF) [55]) have been successfully applied.

Recently, the deep learning method has notably advanced to address challenging problems, such as natural language processing [3, 37, 54], speech and image recognition [12, 18, 20, 30, 67], and computational biology [14, 34–36, 44]. In addition, successful results of deep learning in drug fields [11, 38, 45, 58] have been achieved and attracted further attention. Deep learning approaches to drug analyses can be categorized in the following three cases (see Figure 1):

(A) **Deep classifier:** Deep learning is used as a back-end classifier for manually crafted chemical features. The deep learning based classifier has attracted considerable attention through the Kaggle 'Merck Molecular Activity Challenge' and its subsequent quantitative structure activity relationship (QSAR) predictions [11, 45]. The Kaggle competition and its subsequent predictions showed significant improvements in performance, by using the multi-task technique based on deep learning from the provided features. In addition, prediction of chemical-protein interaction [58] (for drug discovery, network pharmacology, and drug/protein target identification) showed boosted results by using deep learning from manually crafted chemical features. Although deep learning improves prediction performance, it has been used only as a classifier based on the well-crafted chemical features.

(B) **Deep generator:** Deep learning is used as a front-end generator for drug generation. Drug generation extracts the hidden features from the chemicals and then regenerates the chemicals from the extracted features. Generative deep models, such as a variational autoencoder [29] and a recurrent neural network [21], have been used for drug generation [17, 49]. Both models use the simplified molecular input line entry system (SMILES) [57] as input to extract the features and regenerate the SMILES string (or functionally similar SMILES string). Deep learning is only used as a deep generator between chemical strings and extracted features.

(C) **End-to-end deep learning:** Deep learning is used as an end-to-end learner which involves the entire process from feature engineering to predicting biological activity. It should extract hidden features from the original chemical inputs and then predict the biological activity by using the extracted features. However, owing to the considerable performance of using the crafted features and the difficulties of handling the original chemical input, the end-to-end learning framework has not been actively pursued.

Deep learning in drug (chemical) analyses is mainly used as a classifier; however, the original purpose of deep learning is not only for classification, but also for extracting hidden representations [39]. The hidden representations, extracted by deep learning, may have potentially important features that are unknown to domain experts. Therefore, we expect that end-to-end deep learning can improve prediction performance. For this reason, we propose an end-to-end deep learning framework, DeepCCI, for CCI prediction.

## 2 BACKGROUND
## 2.1 Representation of chemical compound

For end-to-end learning, it is important that the input contains all the latent feature information of the chemical compound. A considerable amount of chemical information exists, such as weight, molecular formula, rings, atoms, SMILES [65], and InChI [19]. Among the many pieces of information, SMILES represents the chemical structure as a line of ASCII characters. As shown in Figure 2, cyclohexane and acetaminophen are expressed as C1CCCCC1 and CC(=O)NC1=CC=C(O)C=C1 respectively (C for carbon, O for oxygen, = for double bonds and 1 for the first ring). Atoms (*e.g.*, carbon, nitrogen, and oxygen), bonds (*e.g.*, single, double, and triple bonds), rings (*e.g.*, open ring, close ring, and ring number), aromaticitiy, and branching can be represented with SMILES. As a one-dimensional representation for the chemical structure, SMILES can be converted into a two- or three-dimensional chemical structure, which means that it contains sufficient structure information to derive higher dimensions. SMILES is also used for drug generation [17, 49] and for compound similarity [43].

It is generally known that structure is closely related to function [46, 53]. QSAR prediction, which exploits the relationship between the chemical structure and the biological activity, for identifying 'druglikeness'[1] and establishing metabolic pathways [27, 55]. As an expression of structure information, SMILES can be a means of clarifying the function of a chemical compound.

For these reasons, we used SMILES as a front-end input format for representing a chemical compound.

## 2.2 Convolutional neural networks

The convolutional neural network (CNN) [30], one of the most widely used deep learning architectures, has shown outstanding performance in one-dimensional biological sequences [1, 68] and linguistic sentences [25, 26], as well as in two-dimensional image processing [30, 32, 42].

Designed to analyze shift invariant spatial information, CNNs consist of convolution layers and pooling layers. At each convolution layer, learnable feature maps called filters scan the sub-regions across the sequence. Filters enable CNNs to discover locally correlated motifs regardless of their locations through local connectivity and parameter sharing. Then, each pooling layer summarizes non-overlapping sub-regions, and aggregates local features into more global features. We applied CNNs to capture the SMILES string specificity.

## 2.3 Commutative property

In mathematics, a commutative property means that changing the order does not affect the result (for example, 2+5=5+2).

As in indirect symmetric problems, such as distance or similarity between objects, predicting the interaction between objects A and B is independent of the ordering, namely $\mathcal{I}(A, B) = \mathcal{I}(B, A)$.

However, if the values $\mathbf{x}^A$ and $\mathbf{x}^B$ for objects A and B are simply concatenated, and then used for learning, the result can be different according to the input order

$$\mathcal{I}(\mathbf{x}^A, \mathbf{x}^B) = w_1 x_1^A + \cdots + w_n x_n^A + w_{n+1} x_1^B + \cdots + w_{2n} x_n^B$$
$$\neq w_1 x_1^B + \cdots + w_n x_n^B + w_{n+1} x_1^A + \cdots + w_{2n} x_n^A = \mathcal{I}(\mathbf{x}^B, \mathbf{x}^A). \quad (1)$$

With this difference in mind, we designed a model that produces the same interaction probability regardless of the input order.

## 3 PROPOSED METHODOLOGY

Figure 3 shows the overview of the proposed DeepCCI. Our method presents end-to-end SMILES learning for CCI. It takes SMILES strings as inputs $\mathbf{x}^A$ and $\mathbf{x}^B$ for objects A and B, and then produces an interaction probability $\hat{y}$. The structure is divided into three stages. The first stage is for preprocessing SMILES inputs, the

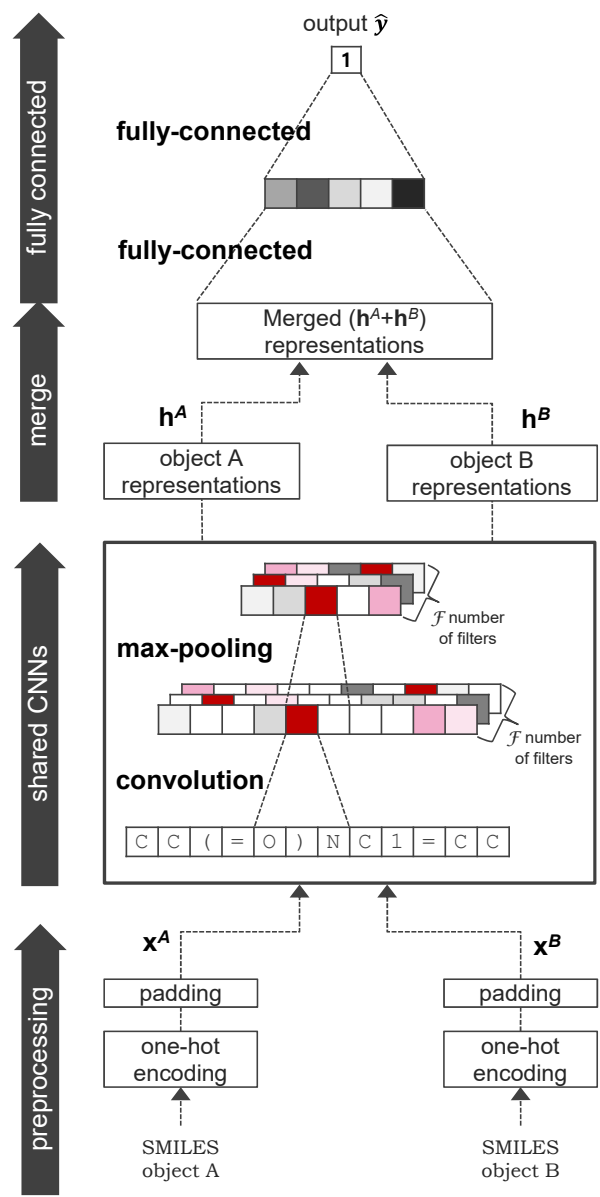

Figure 3: Overview of the proposed DeepCCI end-to-end methodology. Our method extracts hidden representations from SMILES learning and predicts the interaction through the extracted representations, using CNNs and fully connected layers. Inputs are one-hot encoded and zero padded (or truncated) to the maximum length. The preprocessed inputs are fed into shared 1D-CNNs for transforming into hidden representations. The transformed hidden representations for objects A and B are merged and then fed into two fully-connected layers to predict the interaction. The output is the predicted probability of the interaction between objects A and B.

---
[1]Druglikeness is a qualitative indicator to evaluate a drug-like molecule in terms of its bioavailability.

**Algorithm 1** Pseudo-code of proposed DeepCCI

1: **Input:** $\mathbf{x}^A = \left(x_1^A, x_2^A, \cdots, x_\lambda^A\right)$ and $\mathbf{x}^B = \left(x_1^B, x_2^B, \cdots, x_\lambda^B\right)$ ▷ $\mathbf{x}^A, \mathbf{x}^B$: the encoded and padded/truncated inputs for objects A and B
2: **Output:** $\hat{\mathbf{y}}$ ▷ $\hat{\mathbf{y}}$: predicted probability of interaction
3: **repeat**
4:     $\mathbf{h} = \text{Pool}(\sigma_r(\text{Conv}(\mathbf{x})))$ ▷ shared CNNs described in Section 3.3
5:     $\hat{y} = \sigma_s(\text{Fcl}(\sigma_r(\text{Fcl}(\mathbf{h}^A + \mathbf{h}^B))))$ ▷ interaction prediction described in Section 3.4
    ▷ $\mathbf{h}^A, \mathbf{h}^B$: the outputs of shared CNNs from inputs $\mathbf{x}^A, \mathbf{x}^B$
6:     $\mathcal{L}(y, \hat{y}) = -y \log \hat{y} - (1 - y) \log(1 - \hat{y})$ ▷ learning objective to minimize the loss $\mathcal{L}(y, \hat{y})$
7:     $\mathbf{W} = \mathbf{W} - \Delta \mathbf{W}$ ▷ weights update $\Delta \mathbf{W}$ based on averaged $\mathcal{L}$ from mini-batch
8: **until** number of epoch reaches $n_{\text{epoch}}$

second stage is for learning latent hidden representations through 1D-CNNs, and the third stage is for interaction learning through merging and fully-connected layers. Algorithm 1 shows the overall procedure of DeepCCI.

### 3.1 Notations
The notations used in our paper are outlined below.
- $\mathcal{I}(A, B)$: interaction between objects A and B
- $\mathcal{X}$: SMILES character set, $\mathcal{X} = \{\text{C}, =, (, ), \text{O}, \text{F}, 1, 2, \cdots\}$
- $\mathbf{x}$: SMILES string, $\mathbf{x}^A$ and $\mathbf{x}^B$ for objects A and B
- $x_i$: $i$-th character of $\mathbf{x}$, $\mathbf{x} = (x_1, x_2, \cdots, x_n)$, $x_i \in \mathcal{X}$
- $\mathbf{h}$: hidden representations, $\mathbf{h}^A$ and $\mathbf{h}^B$ for objects A and B
- $h_i$: $i$-th element of $\mathbf{h}$, $\mathbf{h} = (h_1, h_2, \cdots, h_n)$
- $\lambda$: maximum length of SMILES
- $\mathcal{F}$: number of filters
- $k$: length of filter (kernel size)
- $\mathbf{W}$: learning weights
- $\hat{y}$: predicted probability of interaction
- $\sigma$: activation function, $\sigma_s$ for the sigmoid, $\sigma_r$ for the rectified linear unit
- $N$: sample size, $N_t$ for total sample, $N_b$ for mini-batch size

### 3.2 Input Preprocessing
We used input $\mathbf{x} = (x_1, x_2, \cdots, x_n)$ which is represented by 65-character $x_i \in \mathcal{X}$ where $\mathcal{X} = \{\text{C}, =, (, ), \text{O}, \text{F}, 1, 2, \cdots\}$, $|\mathcal{X}| = 65$ in the SMILES format. Input $\mathbf{x}$ should be converted into a numeric expression for learning.

For the numerical expression, we employ a widely used one-hot encoding scheme, which sets the corresponding single character to '1' and all the others to '0', thereby converting each character $x_i$ into an $|\mathcal{X}|$-dimensional binary vector. An example of cyclohexane (C1CCCCC1) encoded $\mathbf{x}$ is as follows:

$$\{\text{C}, =, ), (, \text{O}, \text{N}, 1, 2 \cdots \cdots \cdots\}$$

$$\mathbf{x} = \begin{bmatrix} x_1 \\ x_2 \\ x_3 \\ x_4 \\ x_5 \\ x_6 \\ x_7 \\ x_8 \end{bmatrix} = \begin{bmatrix} \text{C} \\ 1 \\ \text{C} \\ \text{C} \\ \text{C} \\ \text{C} \\ \text{C} \\ 1 \end{bmatrix} = \begin{bmatrix} [1, 0, 0, 0, 0, 0, 0, \cdots, 0, 0, 0] \\ [0, 0, 0, 0, 0, 0, 1, 0, \cdots, 0, 0, 0] \\ [1, 0, 0, 0, 0, 0, 0, \cdots, 0, 0, 0] \\ [1, 0, 0, 0, 0, 0, 0, \cdots, 0, 0, 0] \\ [1, 0, 0, 0, 0, 0, 0, \cdots, 0, 0, 0] \\ [1, 0, 0, 0, 0, 0, 0, \cdots, 0, 0, 0] \\ [1, 0, 0, 0, 0, 0, 0, \cdots, 0, 0, 0] \\ [0, 0, 0, 0, 0, 0, 1, 0, \cdots, 0, 0, 0] \end{bmatrix}$$

In general, SMILES strings are variable lengths depending on the complexity of the chemical structure. Chemical compounds of a complex structure have relatively longer SMILES strings compared to compounds of a simple structure. To effectively learn the hidden representations for SMILES, we limit the inputs to a certain maximum length, $\lambda$. If a sequence is shorter than the maximum length, zero values are pre-padded; otherwise, the sequence is truncated to the maximum length. After the padded or truncated processes, all input lengths are fixed to the maximum length, $\lambda$. According to the maximum length, the prediction performance is affected (see Figure 6A).

According to the encoding scheme and maximum length, input $\mathbf{x}$ is encoded into a $\lambda \times |\mathcal{X}|$-dimensional binary matrix. The input collection is represented by a $N_t \times \lambda \times |\mathcal{X}|$-dimensional binary tensor.

### 3.3 Modeling of Hidden Representations
To learn hidden representations $\mathbf{h}$ from preprocessed input $\mathbf{x}$ of SMILES strings, we used CNNs. The CNN architecture, used in our method, consists of a convolution layer and a pooling layer.

$$\mathbf{h} = \text{Pool}(\sigma_r(\text{Conv}(\mathbf{x}))) \quad (2)$$

The convolution layer plays a 'motif detector' role across the sequence. The preprocessed input $\mathbf{x}$ is convoluted with a set of learnable feature maps called filters (or kernels). Different filters may detect motifs of different chemical properties in the SMILES string. Scanning the local region across the input string with a parameter-shared filter may enable recognition of the shift-invariant local pattern.

The output shape of convoluted input $\mathbf{x}$ becomes an $\mathcal{F} \times (\lambda - k + 1)$-dimensional matrix, where $\mathcal{F}$ represents the number of filters, and $k$ represents the filter length.

$$\begin{aligned} \text{Conv}(\mathbf{x}) &= (c_{f,i}) \in \mathbb{R}^{\mathcal{F} \times (\lambda - k + 1)} \\ c_{f,i} &= \sum_{j=1}^{k} \sum_{s=1}^{|\mathcal{X}|} W_{f,j,s} x_{i+j,s} \end{aligned} \quad (3)$$

where $f$ and $i$ are $1 \leq f \leq \mathcal{F}$ and $1 \leq i \leq (\lambda - k + 1)$, respectively. The parameters $\mathcal{F}$ and $k$ of the CNN filter are decided through experiments (see Figure 6D and 6E).

After the convolution layer, each convoluted element is processed with the rectified linear unit (ReLU) [40]

$$\sigma_r(t) = max(0, t) \quad (4)$$

which is an activation function of non-saturating for positives and clamping to zero for negatives.

The pooling layer partitions the rectified responses with non-overlapping sub-regions and then outputs the maximum if it is max-pooling. It summarizes within the sub-regions and reduces the spatial size. The $\mathcal{F}$ dimension remains unchanged. The max-pooling length of 6 showed the best performance in our experiments (see Figure 6C)

To prevent overfitting and to produce generalization effects, we apply dropout [52] for the last CNN units.

For a symmetric problem, such as a tweets comparison, Keras use shared layers, which reuse the weights [9]. Likewise, we use shared CNNs to extract the hidden representations $h^A$ and $h^B$ for chemicals A and B from inputs $\mathbf{x}^A$ and $\mathbf{x}^B$.

## 3.4 Modeling of Interaction Between Chemicals

Unlike heterogeneous interactions, such as protein-chemical and miRNA-mRNA interaction, CCI is a homogeneous interaction between two chemicals. Homogeneous interaction between objects A and B should be the same regardless of their input order, $\mathcal{I}(A, B) = \mathcal{I}(B, A)$, thereby guaranteeing the commutative property described in Section 2.3. However, simple concatenating of the values of objects A and B cannot guarantee the property according to Equation 1.

To equally recognize $\mathcal{I}(A, B)$ and $\mathcal{I}(B, A)$, we strive to give the same weights $\mathbf{W}$ to both hidden representations $\mathbf{h}^A$ and $\mathbf{h}^B$ for objects A and B

$$\begin{aligned}\mathcal{I}(A, B) &\approx \mathbf{W}\mathbf{h}^A + \mathbf{W}\mathbf{h}^B \\ &= w_1 h_1^A + \cdots + w_n h_n^A + w_1 h_1^B + \cdots + w_n h_n^B \\ &= w_1 h_1^B + \cdots + w_n h_n^B + w_1 h_1^A + \cdots + w_n h_n^A \\ &= \mathbf{W}\mathbf{h}^B + \mathbf{W}\mathbf{h}^A \approx \mathcal{I}(B, A). \end{aligned} \quad (5)$$

Through the weight-sharing, we can guarantee the commutative property. The weight-sharing can be implemented by merging both hidden representations through summing process

$$\begin{aligned}\mathbf{W}\mathbf{h}^A + \mathbf{W}\mathbf{h}^B &= w_1 h_1^A + \cdots + w_n h_n^A + w_1 h_1^B + \cdots + w_n h_n^B \\ &= w_1(h_1^A + h_1^B) + \cdots + w_n(h_n^A + h_n^B) \\ &= \mathbf{W}\left(\mathbf{h}^A + \mathbf{h}^B\right). \end{aligned} \quad (6)$$

The interaction between objects A and B, is learned with two fully-connected layers (Fcl) from summed representations.

$$\hat{y} = \sigma_s(\mathbf{Fcl}(\sigma_r(\mathbf{Fcl}(\mathbf{h}^A + \mathbf{h}^B)))) \quad (7)$$

where $\hat{y}$ is the predicted probability that an interaction will occur, $\sigma_r$ is the ReLU in Equation (4) and $\sigma_s$ is the sigmoid activation function

$$\sigma_s(t) = 1/(1 + \exp(-t)) \quad (8)$$

which make the output probability occur from zero to one.

The learning processes proceed to minimize the loss between the true label and the prediction $\mathcal{L}(y, \hat{y})$, where $y$ is the true label and $\hat{y}$ is the predicted probability from Equation (7). For the loss function as the learning objective, we use the binary cross entropy

$$\mathcal{L}(y, \hat{y}) = -y \log \hat{y} - (1 - y) \log(1 - \hat{y}). \quad (9)$$

Table 1: Details of the datasets used in the experiments. The negative samples were randomly generated from negative-0 to balance them with the number of positives

| type/dataset | CCI900 | CCI800 | CCI700 |
| --- | --- | --- | --- |
| # of positives | 9,812 | 75,908 | 171,665 |
| # of negatives | 9,812 | 75,908 | 171,665 |
| total sample size | 19,624 | 151,816 | 343,330 |

Weights are updated in mini-batch units; the loss for the batch unit is averaged by the mini-batch size.

$$\mathcal{L}(\mathbf{y}, \hat{\mathbf{y}}) = \frac{1}{N_b} \sum_{i=1}^{N_b} \mathcal{L}(y_i, \hat{y}_i) \quad (10)$$

where $N_b$ is the mini-batch size, and $i$ represents the sample index.

In the fully-connected layers, we use 128 hidden nodes, batch normalization [23], and a dropout [52] rate of 0.25 for regularization. In our experiments, the entire learning was conducted with the Adam optimizer [28] (learning rate: 0.001, epoch: 100, and mini-batch size: 256). Most of these parameters and network structure were experimentally determined. An epoch of 100 showed stable convergence. We did not use early stopping.

## 4 EXPERIMENTAL RESULTS
### 4.1 Experimental Setup

*4.1.1 Dataset.* In our experiment, we used the CCI data (version 5.0) downloaded from the STITCH [31] database. From the CCI data, only interaction pairs composed of a single compound were used, and the combined compounds were removed. Each interaction pair had a text mining score ranging from 0 to 999, The higher the score was, the greater the interaction probability was. The positive-900 (9,812 samples), positive-800 (75,908 samples), and positive-700 (171,665 samples) were generated from the samples with scores more than 900, 800, and 700, respectively. The negative-0 (212,741 samples) were generated from the samples with scores of zero.

According to the score, three datasets were prepared: chemical-chemical interaction over 900 (CCI900), 800 (CCI800), and 700 (CCI700) as listed in Table 1. The total number of the positive-900, 800, and 700 samples were included in the CCI900, CCI800, and CCI700, respectively. The negatives were randomly selected from negative-zero (212,741) to balance them with the number of positives.

All three datasets were asymmetric, which means that only one each of the A-B pair and B-A pair was included; its opposite pair was not included in the dataset. Since methods of supporting the commutative property are advantageous to the symmetric dataset, we prepared the asymmetric dataset for a fair comparison.

*4.1.2 Preparation of Input Representation.* The representations of chemical compounds were prepared as two types: the SMILES string, and PubChem fingerprint [64]. The PubChem fingerprint is widely used descriptor. It has showed the best performance for certain tasks [48, 63], although the performance of the molecular descriptor depends on the target problem and classifier.

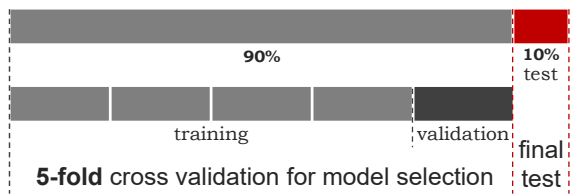

**Figure 4: Dataset breakdown.** The full dataset was randomly divided into the model selection set (90%) and final test set (10%). The model selection set was again partitioned into five sets, for 5-fold cross validation. Through the iterative validation, the model with hyperparameters, showing the best average performance, was selected.

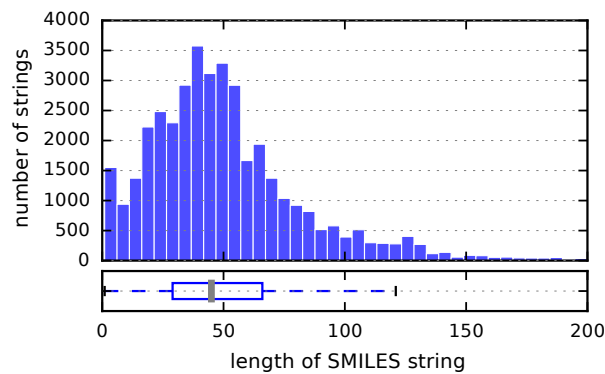

**Figure 5: SMILES string length distribution based on CCI900:** the median length is 45 and the average length is 53.5. The 91% of SMILES strings is less than 100 characters in length.

Strings represented in SMILES were used as inputs for the proposed end-to-end deep learning method. Binary feature vectors represented in the PubChem fingerprint were used as inputs for the other methods. We prepared the PubChem fingerprint accordingly because it is difficult to handle the string in conventional machine learning methods, and the end-to-end learning performance had to be compared with a plain deep classifier (see Figure 1A).

The PubChem fingerprint represents the chemical properties as a 881-dimensional binary vector that has feature information, such as the presence or count of atoms, rings, and nearest neighbor patterns. We concatenated the fingerprints of two objects as a (2×881)-dimensional binary vector and then used it as the inputs for the learning algorithms tested.

From the chemical IDs listed in the STITCH database, we retrieved the SMILES string and PubChem fingerprint by using PubChemPy [56]. A representative example of cyclohexane is as follows:

- PubChem chemical ID: 8078
- SMILES string: C1CCCCC1
- Fingerprint: 1100000001100000000000000000000000000··· (the 9-th bit is 1 if there are more than two carbon atoms and is 0 otherwise; the 10-th bit is 1 if there are more than four carbon atoms and is 0 otherwise)

*4.1.3 Experimental Configuration.* Our dataset consisted of training, validation, and test samples, as in the general learning experiments [2]. Figure 4 details the dataset. The full dataset was divided into two parts, one for model selection and the other for the independent test.

The goal of model selection was to find the optimal hyperparameters. Models with various sets of hyperparameters were trained and then iteratively validated through $k$-fold cross validation. Optimal hyperparameters, which showed the best average performance in terms of AUC, were selected. In our experiments, we used 5-fold cross validation.

The remaining test set was used to assess the generalization of the selected model, and compare its performance with other learning methods. Since the parameters were determined through training and validation independently of the test set, it could guarantee the fairness of the independent test.

Our experiments were run on Ubuntu 14.04 (3.5GHz Intel i7-5930K and GTX Titan X Maxwell (12GB)). For the implementation of deep learning, we used the Keras library package (version 1.1.1) [9].

*4.1.4 Evaluation Metrics.* For the performance comparison, we used seven different evaluation metrics:

- AUC: the area under the ROC curve, which is widely used in case of binary classification to show the overall performance of a binary classifier[2]
- ACC: accuracy, which is intuitively used metric to assess the performance[3]
- TPR: true positive rate, sensitivity or recall[4]
- TNR: true negative rate or specificity[5]
- PPV: positive predictive value or precision[6]
- NPV: negative predictive value[7]
- F1: F1 score[8]

## 4.2 Effects of Hyperparameter Variation

Five different hyperparameter variation experiments were performed to find the empirical optimal parameters.

As shown in Figure 5, the lengths of SMILES strings vary from 1 to over 200. Their median length is 45 and the average length is 53.5. The 84%, 91%, and 94% of SMILES strings are less than 80, 100, and 120 characters in length, respectively. The proposed method DeepCCI limits the SMILES strings to a certain maximum length, $\lambda$. The loss of information and dummy zeros according to the maximum length might affect the overall performance. Figure 6A depicts the performance of DeepCCI by varying the maximum

---

[2]The area under the ROC curve which plots $TPR(T)$ versus $FPR(T)$ with varying threshold $T$, where $FPR = 1 - TPR$
[3]$ACC = (TP + TN)/(TP + TN + FP + FN)$, where
TP: true positive, FP: false positive, FN: false negative, TN: true negative
[4]$TPR = TP/(TP + FN)$
[5]$TNR = TN/(TN + FP)$
[6]$PPV = TP/(TP + FP)$
[7]$NPV = TN/(TN + FN)$
[8]$F1 = 2TP/(2TP + FP + FN)$

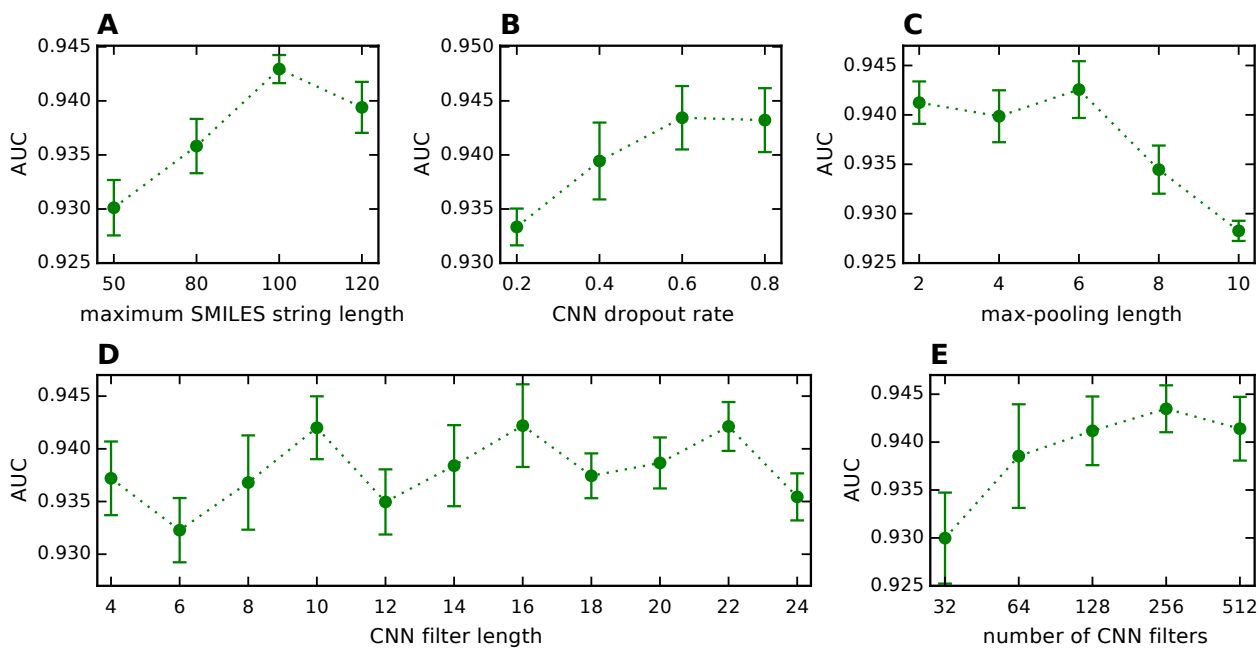

Figure 6: Hyperparameter optimization using 5-fold cross validation tested on CCI900. The baseline hyperparameters of the maximum SMILES string length, CNN dropout rate, max-pooling length, CNN filter length, and number of CNN filters are 100, 0.6, 6, 22, and 256, respectively. The performance changes as varying the (A) maximum SMILES string length from 50 to 120, (B) CNN dropout rate from 0.2 to 0.8, (C) max-pooling length from 2 to 10, (D) CNN filter length from 4 to 24, and (E) number of CNN filters from 32 to 512.

SMILES string length to 50, 80, 100 and 120. Among them, a maximum length of 100 shows the best performance in terms of AUC. In our experiments, the maximum length of 100 could mean that learning was achieved by reducing the computational burden while covering as much information as possible.

Figure 6B illustrates the performance as varying the CNN dropout rate from 0.2 to 0.8. Dropout as a regularization technique reduces overfitting by dropping out units in neural networks. In our validation experiments, a dropout rate of 0.6, meaning that 60% of units dropped out, showed the best performance and the lowest standard deviation.

The pooling layer summarized the adjacent features resulting abstract representations, and inputs were down-sampled, resulting in a smaller number of model parameters to learn [2]. A larger pooling length means a greater summary effect. In our experiments, pooling lengths of 2 and 6 showed similar performances in terms of AUC, as shown in Figure 6C. To achieve better summarization effects, we used the max-pooling with a length of 6 as the baseline parameter.

The key features of CNNs are a parameter sharing architecture and translation invariance characteristics, which are due to the use of filters (or kernels). The filter length depends on the characteristic of patterns to discover. The number of filters controls the capacity, depending on the complexity of the task. As shown in Figure 6D and 6E, we used the filter length of 22 and 256 filters as the baseline parameters. The filter lengths of 10, 16, and 22 (6 units apart) showed the highest AUC. The relatively longer molecule patterns compared to those of the PubChem fingerprint can be captured with these CNN filters. A PubChem fingerprint takes complicated neighbor patterns into consideration, but most of its features are targeted for short and simple patterns.

### 4.3 Performance Comparison with Other Methods

We compared our proposed method, DeepCCI, with a feedforward neural network (FFNN), which is a plain deep classifier described in Figure 1A, and the conventional machine learning methods of support vector machine (SVM) [61, 62], random forest (RF) [5, 55], and adaptive boosting (AdaBoost) [16]. SVM, RF, and AdaBoost were tested with default parameters. FFNN, which is a deep neural networks of three fully-connected layers (with 1,024 and 128 hidden units) was tested with ReLU for the hidden activation function and sigmoid for the output activation function, and the dropout rate of 0.25.

In the context of drug development, the high false positive rate means that there are more candidate chemicals to be screened, while the high false negative rate means that more chemicals, which should not be omitted in the screening list, are omitted. Therefore, both FP and FN rates should be properly controlled. We used seven evaluation metrics related to false positives and false negatives.

Table 2: Details of experimental results based on seven different evaluation metrics were obtained. The experiments were carried out on three dataset with five different methods. All values were the average of ten repeated experiments. The performances in terms of AUC and ACC are represented with bar plots in Figure 7.

| data | methods | AUC | ACC | TPR | TNR | PPV | NPV | F1 |
|---|---|---|---|---|---|---|---|---|
| CCI900 | **DeepCCI** | **0.9483** | **0.8819** | **0.8823** | **0.8815** | **0.8770** | **0.8868** | **0.8796** |
|  | FFNN | 0.9313 | 0.8611 | 0.8424 | 0.8790 | 0.8696 | 0.8536 | 0.8557 |
|  | SVM | 0.8399 | 0.7658 | 0.7225 | 0.8073 | 0.7830 | 0.7531 | 0.7509 |
|  | RF | 0.9016 | 0.8217 | 0.8616 | 0.7836 | 0.7922 | 0.8554 | 0.8254 |
|  | AdaBoost | 0.7742 | 0.7076 | 0.6865 | 0.7278 | 0.7071 | 0.7081 | 0.6966 |
| CCI800 | **DeepCCI** | **0.9904** | **0.9562** | **0.9703** | **0.9419** | **0.9439** | **0.9693** | **0.9569** |
|  | FFNN | 0.9776 | 0.9303 | 0.9339 | 0.9269 | 0.9279 | 0.9330 | 0.9308 |
|  | SVM | 0.8711 | 0.7855 | 0.7678 | 0.8034 | 0.7980 | 0.7753 | 0.7820 |
|  | RF | 0.9589 | 0.8918 | 0.9196 | 0.8638 | 0.8717 | 0.9144 | 0.8950 |
|  | AdaBoost | 0.7994 | 0.7311 | 0.7014 | 0.7609 | 0.7470 | 0.7169 | 0.7235 |
| CCI700 | **DeepCCI** | **0.9932** | **0.9634** | **0.9773** | **0.9495** | **0.9507** | **0.9768** | **0.9639** |
|  | FFNN | 0.9862 | 0.9482 | 0.9509 | 0.9455 | 0.9457 | 0.9508 | 0.9483 |
|  | SVM | 0.8808 | 0.7971 | 0.7655 | 0.8287 | 0.8184 | 0.7811 | 0.7899 |
|  | RF | 0.9709 | 0.9134 | 0.9352 | 0.8918 | 0.8961 | 0.9323 | 0.9152 |
|  | AdaBoost | 0.8082 | 0.7317 | 0.7114 | 0.7519 | 0.7410 | 0.7231 | 0.7259 |

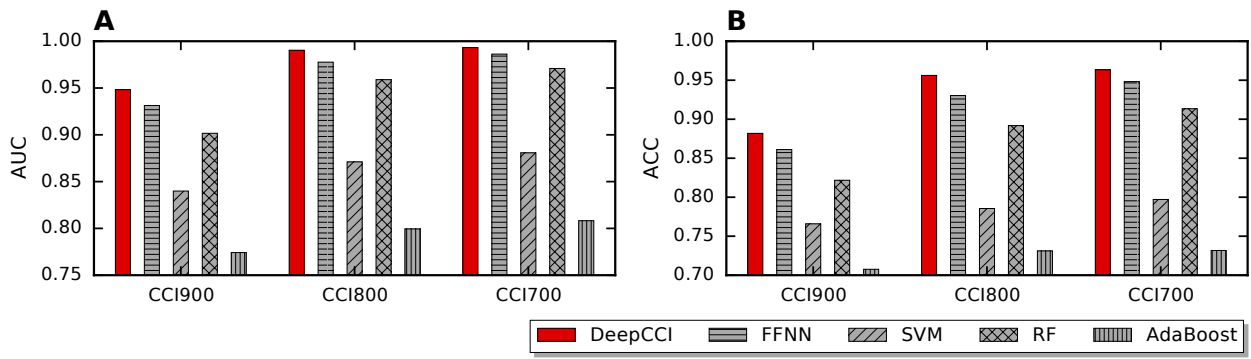

Figure 7: Visualized performance comparison in Table 2 for CCI900, CCI800, and CCI700 in terms of (A) AUC and (B) ACC.

Table 2 shows the overall detailed experimental results in terms of seven evaluation metrics, Figure 7 shows the visualized performance in terms of AUC and ACC with bar plots. All the values were averaged through ten repeated experiments and showed to be stable with a standard deviation of less than 0.004 in terms of AUC for all datasets. In all the metrics and datasets, DeepCCI showed the best performance, followed by FFNN and RF. AdaBoost and SVM showed unsatisfactory results, giving the accuracy of under 0.8 for all datasets. In terms of AUC, DeepCCI, FFNN, and RF were greater than 0.9 for all datasets, and DeepCCI outperformed the FFNN and RF by 1.82% (statistically significantly different, $p$-value[9] of 2.3E-09) and 5.17% ($p$-value of 1.7E-11), respectively for ICC900 dataset. The overall performance gradually increased as the dataset size increased.

The differences in performance between DeepCCI and FFNN could be interpreted as a result of the difference between the end-to-end SMILES and the crafted PubChem fingerprint. The end-to-end feature learning compared to the fingerprint showed better performance. We can surmise that potential key features are automatically extracted, which may be unknown to domain experts.

Figure 8 shows the experimental validation of the commutative property trained with CCI900 training data (order of A followed by B (A,B)) and tested with the switched order CCI900 training data (B followed by A (B,A)). DeepCCI showed the same training and test accuracy, on account of the effect of guaranteeing the commutative property. However, the test accuracy of FFNN, SVM, RF, and AdaBoost showed performance degradation by 18%, 17%, 21%, and 10% compared to the training accuracy. These results mean that the four methods recognized $\mathcal{I}(A,B)$ and $\mathcal{I}(B,A)$ independently and produced different prediction probabilities for them, whereas DeepCCI recognized them as the same.

---
[9]$p$-value from a $t$-test for ten repeated experimental results in terms of AUC of DeepCCI and the other method

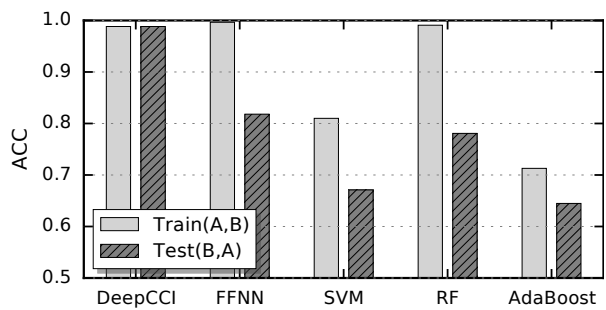

**Figure 8: Experimental validation of the commutative property. Training accuracy for $\mathcal{I}(A,B)$ and test accuracy for switched order $\mathcal{I}(B,A)$.**

## 5 DISCUSSION

In this paper, we proposed the first end-to-end deep learning method for CCI. It showed the best performance compared to a plain deep classifier and conventional machine learning methods.

Deep learning has shown remarkable success in chemical analyses and attracted considerable research attention. Despite the success and interest, the crafted features designed by domain experts continue to be used, even in state-of-the-art deep learning approaches. The objective of deep learning is not merely classification or regression from the crafted features, but automatically extracting meaningful features from the original input and analyzing them. For this reason, we have strived to perform end-to-end deep learning based on SMILES. Our approach showed better results than using the crafted PubChem fingerprint. With the CNN filters used in our experiments, we can capture relatively longer patterns that might otherwise be hard to discover with manually crafted PubChem fingerprints. We believe that patterns longer than 10 characters as well as short and simple patterns in a SMILES string play a particular role in biological reaction between chemicals. There are a variety of well-refined chemical features, and PubChem is one of the competent and widely used descriptors. Through a comparison with the PubChem, we confirmed the possibility of end-to-end SMILES learning, although we did not compare it with all descriptors.

Automatic feature learning alleviates the sophisticated efforts required for feature engineering of domain experts and it can enable discovery of meaningful features that are possibly unknown to domain experts. We expect that the proposed end-to-end SMILES learning is applied to other drug (chemical) analyses, such as QSAR prediction, chemical toxicology, and drug target prediction, and it will improve prediction performance.

In addition to presenting end-to-end SMILES learning framework, we guaranteed the commutative property in the prediction of CCI. Since CCI is a homogeneous interaction problem, it should produce the same interaction probability for $\mathcal{I}(A,B)$ and its converse, $\mathcal{I}(B,A)$. To guarantee the commutative property we used model sharing and merging hidden representations techniques and confirmed the property with experimental verification. We expect this technique can be applied to other homogeneous symmetric problems, such as protein-protein and gene-gene interactions, as well as other homogeneous comparisons.

The future revisions of DeepCCI may include the following improvements: We are planning to analyze how the features automatically extracted from shared CNNs affect the overall performance when they are applied to the conventional machine learning classifiers. As in other fingerprint-based approaches, deep learning-based hidden features may be used as a separate fingerprint for various classifiers. Instead of zero-padding, we may consider using a wildcard character in encoding as a special character, as was done in the idea of wildcard denoising [51]. All the zero values included in one-hot encoding may confuse the network, making it fail to recognize them as a padded region.

## ACKNOWLEDGMENTS

The authors would like to thank the anonymous reviewers of this paper for their constructive comments and helpful suggestions. This work was supported in part by the National Research Foundation of Korea (NRF) grant funded by the Korea government (ICT and Future Planning) [No. 2015M3A9A7029735], in part by a grant of the Korea Health Technology R&D Project through the Korea Health Industry Development Institute (KHIDI), funded by the Ministry of Health & Welfare [HI14C3405030014 and HI15C3224] and in part by the Brain Korea 21 Plus Project in 2017.